\documentclass[letterpaper, 10 pt, conference]{ieeeconf}  

\IEEEoverridecommandlockouts                              

\usepackage{graphicx}
\usepackage{mathtools}
\usepackage{amssymb}
\usepackage{amsmath}
\usepackage{array}
\usepackage{algorithm}
\usepackage{algorithmic}
\usepackage{tabularx}
\usepackage{multirow}
\usepackage[noadjust]{cite}
\usepackage{flushend}
\usepackage{url}

\title{\LARGE \bf
End-to-End Motion Planning of Quadrotors Using Deep Reinforcement Learning
}

\author{Efe Camci$^{1}$ and Erdal Kayacan$^{2}$
\thanks{$^{1}$Efe Camci is with School of Mechanical and Aerospace Engineering, Nanyang Technological University, 50 Nanyang Avenue, 639798, Singapore. 
        {\tt\small efe001@e.ntu.edu.sg}}%
\thanks{$^{2}$Erdal Kayacan is with Department of Engineering, Aarhus University, DK-8000 Aarhus C, Denmark. 
        {\tt\small erdal@eng.au.dk}}
}

\begin{document}

\maketitle
\thispagestyle{empty}
\pagestyle{empty}

\begin{abstract}
In this work, a novel, end-to-end motion planning method is proposed for quadrotor navigation in cluttered environments. The proposed method circumvents the explicit sensing-reconstructing-planning in contrast to conventional navigation algorithms. It uses raw depth images obtained from a front-facing camera and directly generates local motion plans in the form of smooth motion primitives that move a quadrotor to a goal by avoiding obstacles. Promising training and testing results are presented in both AirSim simulations and real flights with DJI F330 Quadrotor equipped with Intel RealSense D435. The proposed system in action can be found in \url{https://youtu.be/pYvKhc8wrTM}.
\end{abstract}

\section{INTRODUCTION}  
Autonomous navigation of quadrotors necessitates sensing \cite{huang2017visual,stevens2018vision,sun2018robust,yang2019cubeslam}, planning \cite{camci2019planning,morrell2018comparison,lai2018optimal,camci2019learning}, and control \cite{greeff2018flatness,tal2018accurate,tang2018learning,mehndiratta2018automated}. Separation of these tasks is the medium within the current state-of-the-art navigation methods. Each task is performed by an individual module and modularity is attained easily by this way. Nevertheless, modularity comes with the cost of possible incompatibility, especially with the presence of erroneous modules. An erroneous module in the pipeline could easily cause the other modules to fail as well. Therefore, in this work, the unification of these tasks is attempted within a single, reliable module using deep reinforcement learning (RL) \cite{sampedro2018image,everett2018motion,do2018learning,gschwindt2019can}. 

The proposed method aims to solve the following navigation problem: A quadrotor is deployed for navigation in a partially known environment. A rough path to the goal position is known without any obstacle location information on it. The quadrotor is supposed to generate local motion plans using this information and its online sensory data for safe and quick navigation. To this end, a deep RL agent is proposed which uses raw depth images from a front-facing camera to generate desirable motion primitive sequences. Particularly, a deep Q-network (DQN) with around 75,000 parameters is trained which takes a raw depth image and relative position information as its input, and yields a motion primitive selection as its output.


\section{APPROACH} 
\begin{figure}[t!]
  \centering
  \includegraphics[width=\columnwidth]{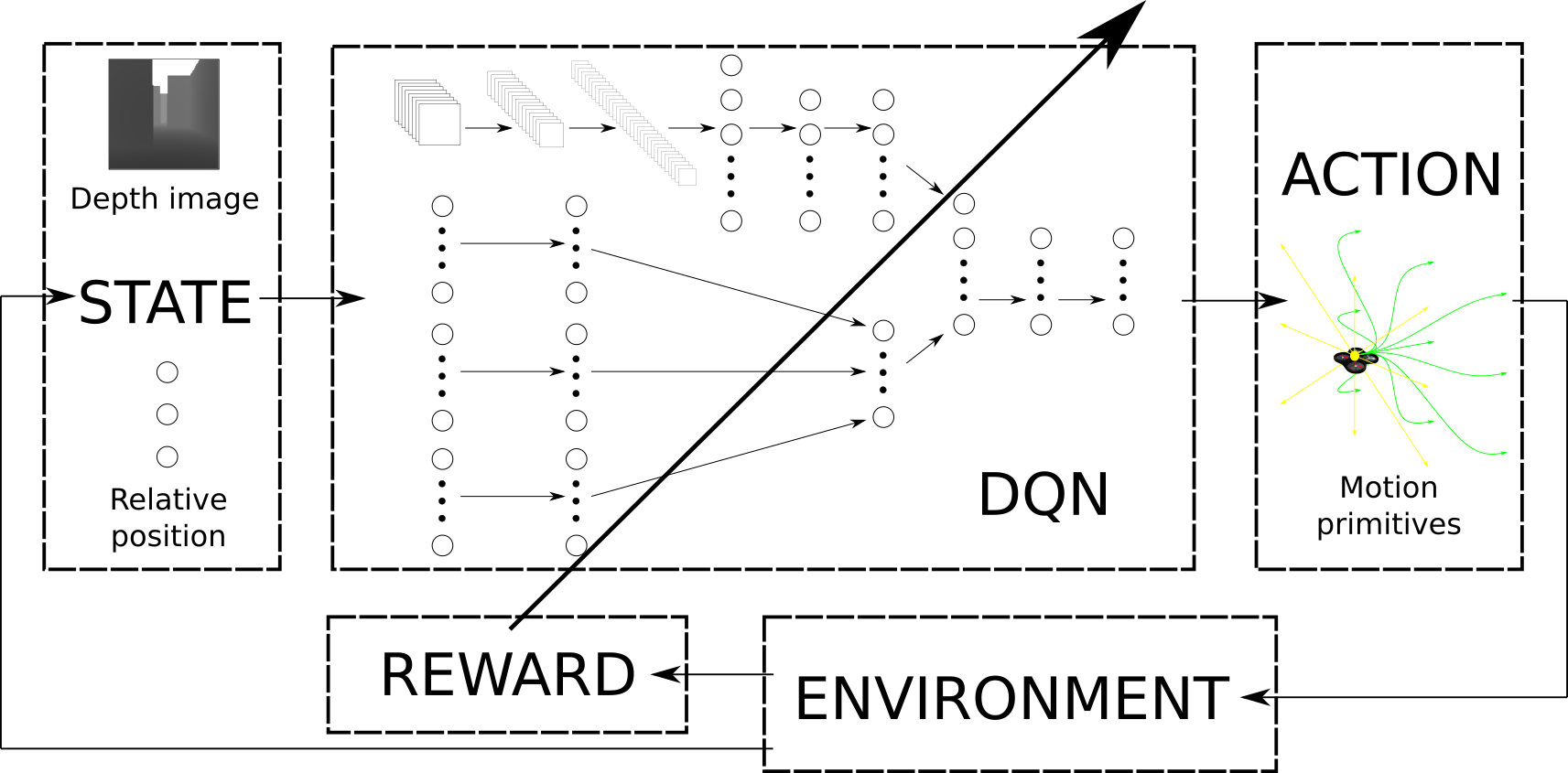}
  \caption{Overview of the proposed RL system.}
  \label{overview}
\end{figure}

In RL, it is critical to design the main elements properly: state, action, and reward. Each of these elements is explained in the following subsections while being depicted as the main elements of the proposed RL system in Fig.~\ref{overview}.

\subsection{State}
State is the respective situation of the agent in the environment. It is designed as a pair of a depth image and relative position information. The depth image has the size of 32$\times$32 and it is obtained by the front-facing camera. It is included in the state definition in order to inform the agent about the obstacles around. The relative position on the other hand is a 3$\times$1 vector. It is calculated by substracting the quadrotor's position vector from the moving setpoint's position vector, and transforming the resultant vector into the body frame of the quadrotor $x_B$, $y_B$, $z_B$. It is included in the state definition in order to inform the agent about its respective motion along the rough path towards the goal.
  
\subsection{Action}
Action is the agent's respective move at each time step following some policy to increase rewards. It is in the form of motion primitives whose basis is formed by B\'{e}zier curves. They are parametric curves based on Bernstein polynomials:
\begin{equation} \label{eq_bezier}
C:[0,1]\longrightarrow\mathbb{R},\;\;\;C(t) = \sum\limits_{i=0}^{n} \binom{n}{i}P_i B_{n,i}(t),
\end{equation}
where $P_i$ are control points and $B_{n,i}(t)$ are Bernstein polynomials of $n^{th}$ degree which are given as:
\begin{equation} \label{eq_bernstein}
B_{n,i}(t)=\left(1-t\right)^{\left(n-i\right)}t^{i}.
\end{equation}

\begin{figure}[t!]
  \centering
  \includegraphics[width=\columnwidth]{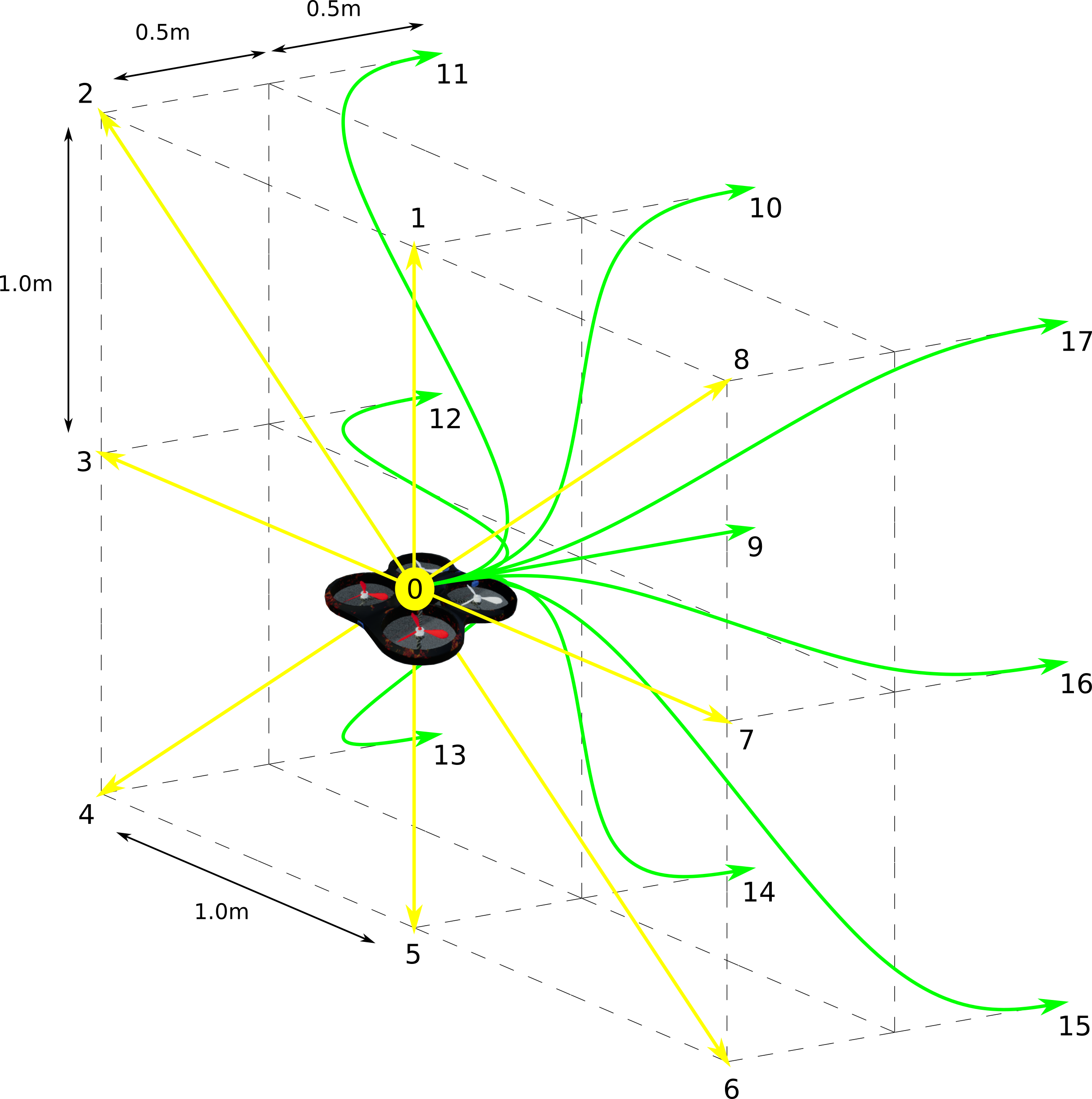}
  \caption{Motion primitives as action set.}
  \label{fig_action_set}
\end{figure}

Smooth motion primitives in the position domain are generated by utilizing the cubic B\'{e}zier curves (n=3) for each finite time step. At each time step, the agent selects an action among the action set which consists of 18 different primitives (Fig.~\ref{fig_action_set}). This set is designed on the account that the agent's motion would be biased towards forward because its only exteroceptive sensor is a front-facing depth camera.

\subsection{Reward}
Reward is the signal which assesses the quality of the agent's actions. It is designed based upon the quadrotor's relative motion with respect to the moving setpoint on the initial rough path. It is equal to the change in the Euclidean distance ($\Delta d$) between the moving setpoint and the quadrotor position through a time step, discounted by the Euclidean distance between the moving setpoint and the quadrotor position at the end of the current time step ($d_t$). The reward function also incorporates a primary logic based on excessive deviations from the initial path and collisions with obstacles. In this vein, higher rewards are obtained for motions close to the initial path. Lower rewards are obtained for motions away from the initial path. Collisions result in a drastic punishment. Excessive deviations from the initial path (the ones that are larger than 5m) result in a milder punishment. This reward logic is designed for the agent to learn obstacle avoidance and quick navigation towards the goal at the same time in unknown environments. It is mathematically defined as:
\begin{equation} \label{eq_reward}
R=
\begin{dcases}
    \frac{R_l}{d_t},               & \text{for}\;\; \Delta d_u < \Delta d \\
    \frac{R_l + \left(R_u-R_l\right)\frac{\Delta d_u-\Delta d}{\Delta d_u-\Delta d_l}}{d_t},   & \text{for}\;\; \Delta d_l \leq \Delta d \leq \Delta d_u \\
    \frac{R_u}{d_t},               & \text{for}\;\; \Delta d < \Delta d_l \\
    R_{dp},               & \text{for excessive deviation,} \\
    R_{cp},               & \text{for collision.}
\end{dcases}
\end{equation}

The variables $R_l$ and $R_u$ are the reward boundaries, and they are 0 and 0.5 respectively. The terms $\Delta d_l$ and $\Delta d_u$ are the reward saturation bounds on $\Delta d$, and they are -1m and 1m, respectively. The term $R_{dp}$ is the excessive deviation punishment and it is equal to -0.5. The term $R_{cp}$ is the collision punishment and it is equal to -1. Most of these parameters are determined by trial-and-error method. In fact, they are flexible to be modified based upon the case study of interest.  
  
\subsection{Algorithmic Details} 
The idea behind many RL algorithms is to estimate the action-value function $Q(s,a)$. This function can get intricate in the case of real robots due to extremely high number of state-action pairs. Therefore, function approximators have emerged as a common choice to estimate action-value function by adding a certain level of generalization \cite{mnih2015human}. In this work, a DQN is utilized for estimating the complex action-value function which is given as \cite{mnih2015human}:
\begin{equation} \label{eq_qfunction}
Q(s,a)=\mathbb{E} \left(R_{t+1}+\gamma \displaystyle \mathop{\max}_{a'} Q(s_{t+1},a') \big| s_t=s, a_t=a \right)
\end{equation}
where $s_t$ is the current state of the agent at the time step $t$. The term $a_t$ is the action taken by the agent. The term $s_{t+1}$ is the next state that the agent reaches after taking the action $a_t$. The term $R_{t+1}$ is the reward that the agent gets as a result of its action. The term $\gamma$ is the discount factor which determines the present value of future rewards \cite{sutton1998reinforcement}. 

\begin{figure*}[t!]
  \centering
  \includegraphics[width=\textwidth]{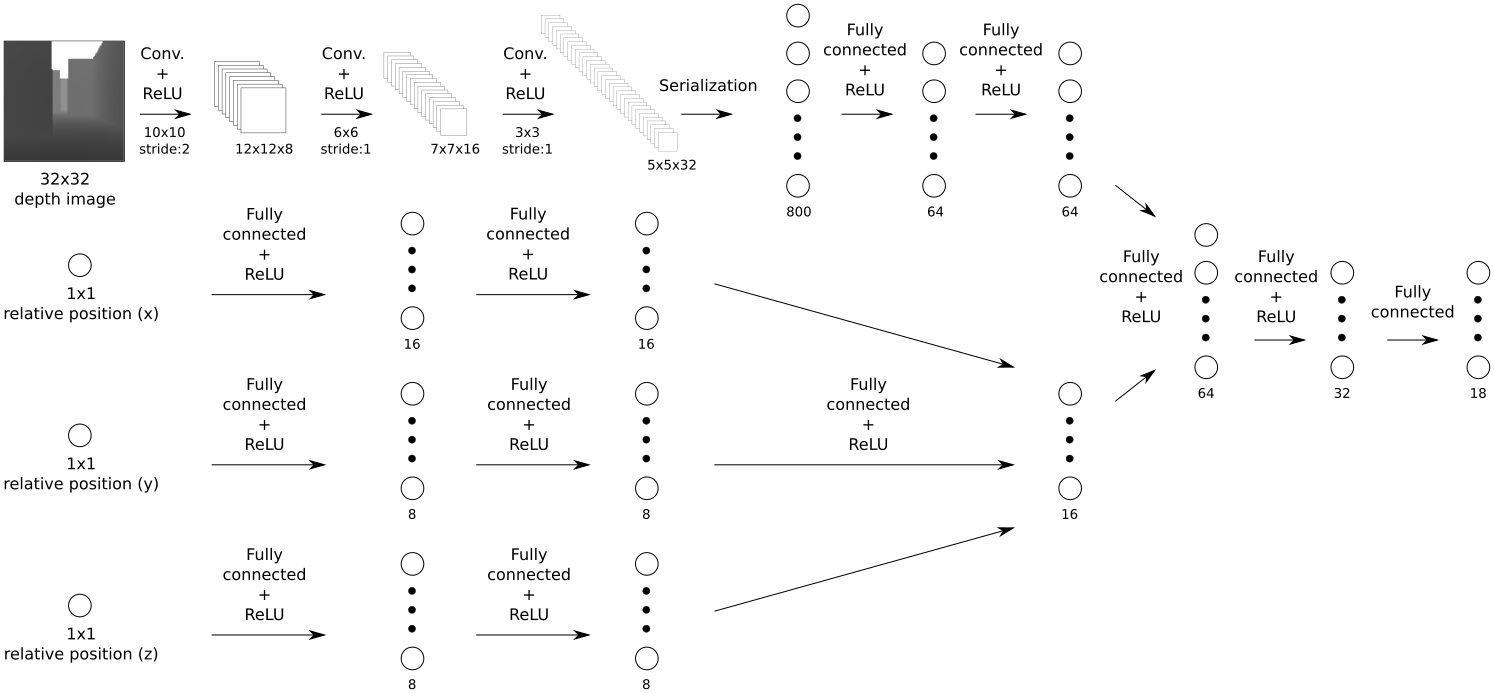}
  \caption{DQN architecture composed of convolutional and fully connected layers.}
  \label{fig_nn_arch}
\end{figure*}

The DQN considered in this work is a combination of convolutional and fully connected neural networks which fuses two different inputs in two main lanes. The first lane uses hierarchical layers of convolutional filters to reason about correlations between local spatial portions of the depth image while the second lane utilizes fully connected layers to incorporate the relative position information. The architecture is depicted in detail in Fig. \ref{fig_nn_arch}. The first convolutional layer in the first lane takes a 32$\times$32 raw depth image as the input and convolves it with 10$\times$10 sized 8 filters with stride 2. It then applies a rectified linear unit (ReLU). The second and third convolutional layers use 16 filters of size 6$\times$6 and 32 filters of size 3$\times$3 with stride 1, respectively. They also apply ReLU after convolution. Subsequently, a fully connected layer of size 800 gets the output of the third convolutional layer and feeds it to the following fully connected layers of sizes 64. In the second lane, the relative position of the moving setpoint with respect to the quadrotor is fed to fully connected layers in three sub-lanes with sizes 16\footnote{The first sub-lane is designed to have a larger layer size as compared to the other two in order to put more emphasis on the relative position information in $x_B$. This information is possibly more important for successful navigation to the goal considering the forward-biased motion of the quadrotor.}, 8, 8. They are then fully connected to a single layer of size 16. Subsequently, the outputs of the first lane (64) and the second lane (16) are concatenated and fed to the final fully connected layers of sizes 64, 32, and 18. All of the fully connected layers in this DQN structure apply ReLU except from the last one which eventually yields the Q-value estimation for each motion primitive. This network is trained using Huber loss \cite{huber1964robust} through Adam optimizer \cite{kingma2014adam} in PyTorch with the default settings.

\section{EXPERIMENTS}
The experiments for validating the method consist of training and testing stages. Firstly, the agent is trained in seven different environments in AirSim in order to have diversified retrospective knowledge. Then, it is tested in these seven environments as well as three other environments which are unseen during the training stage. Lastly, the agent is also deployed for preliminary real flight tests to demonstrate the applicability of the proposed method for real vehicles.

\subsection{Training in AirSim}
\begin{figure}[t!]
  \includegraphics[width=\columnwidth]{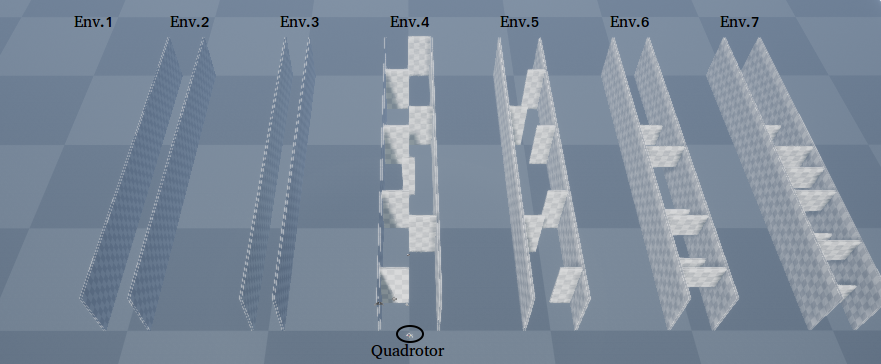}
  \caption{Training environments: obstacle-free, corridor, and slalom tracks.}
  \label{airsim_training}
\end{figure}

The training environments in AirSim are depicted in Fig.~\ref{airsim_training}. The environments are diversified in terms of their complexity, from obstacle-free environment (Env. 1) and obstacle-free corridors with different widths (Env. 2, Env. 3) to left-right (Env. 4, Env. 5) and up-down (Env. 6, Env. 7) slalom environments. All of these environments are merged in a single AirSim session, and they are visited by the agent randomly. Merging them together is particularly helpful because diversification of data samples during learning is improved by this way, enhancing the data sample efficiency. Since the main aim in this work is to develop a simple RL system with relatively early convergence for navigation of quadrotors, high data sample efficiency can be substantially useful. In the same vein, it is also attempted to elude the common need for a large number of interactions in RL because high fidelity AirSim simulations run on real-time. In AirSim, it may require weeks or months to obtain the same amount of data which could have been obtained from accelerated, simple simulation environments in a single day.

The convergence pace in RL depends on many factors such as randomized data, network size, $\varepsilon$, $\gamma$, learning rate of the back-propagation algorithm used for DQN, architecture of DQN, etc. For the sake of brevity, only a single set of hyperparameters is considered which is observed to be working well throughout the training trials during this work. The minimum number of episodes required for convergence is found out by fixing the network structure, $\varepsilon$, and $\gamma$ for five training sessions which consist of 100, 200, 500, 1000, and 2000 episodes. In each session, a linear $\varepsilon$ decay from 1.0 to 0.1 and a linear $\gamma$ gain from 0.01 to 0.99 for the first 80\% of total episodes are utilized, while these parameters are kept constant for the next 20\%.

\begin{figure}[t!]
  \centering
  \begin{minipage}{0.49\columnwidth}
    \centering
    \includegraphics[width=\columnwidth]{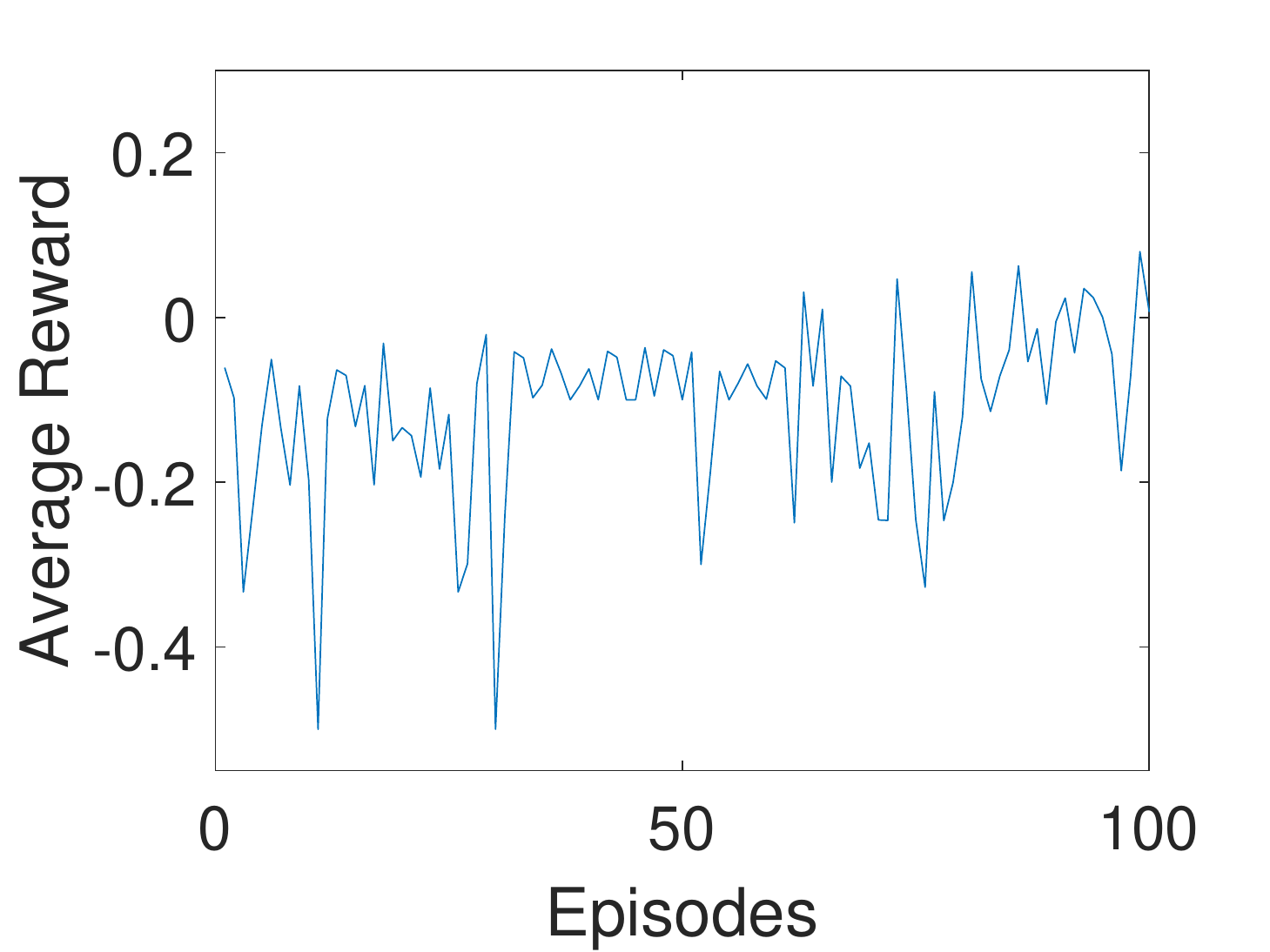}
  \end{minipage}
  \begin{minipage}{0.49\columnwidth}
    \centering
    \includegraphics[width=\columnwidth]{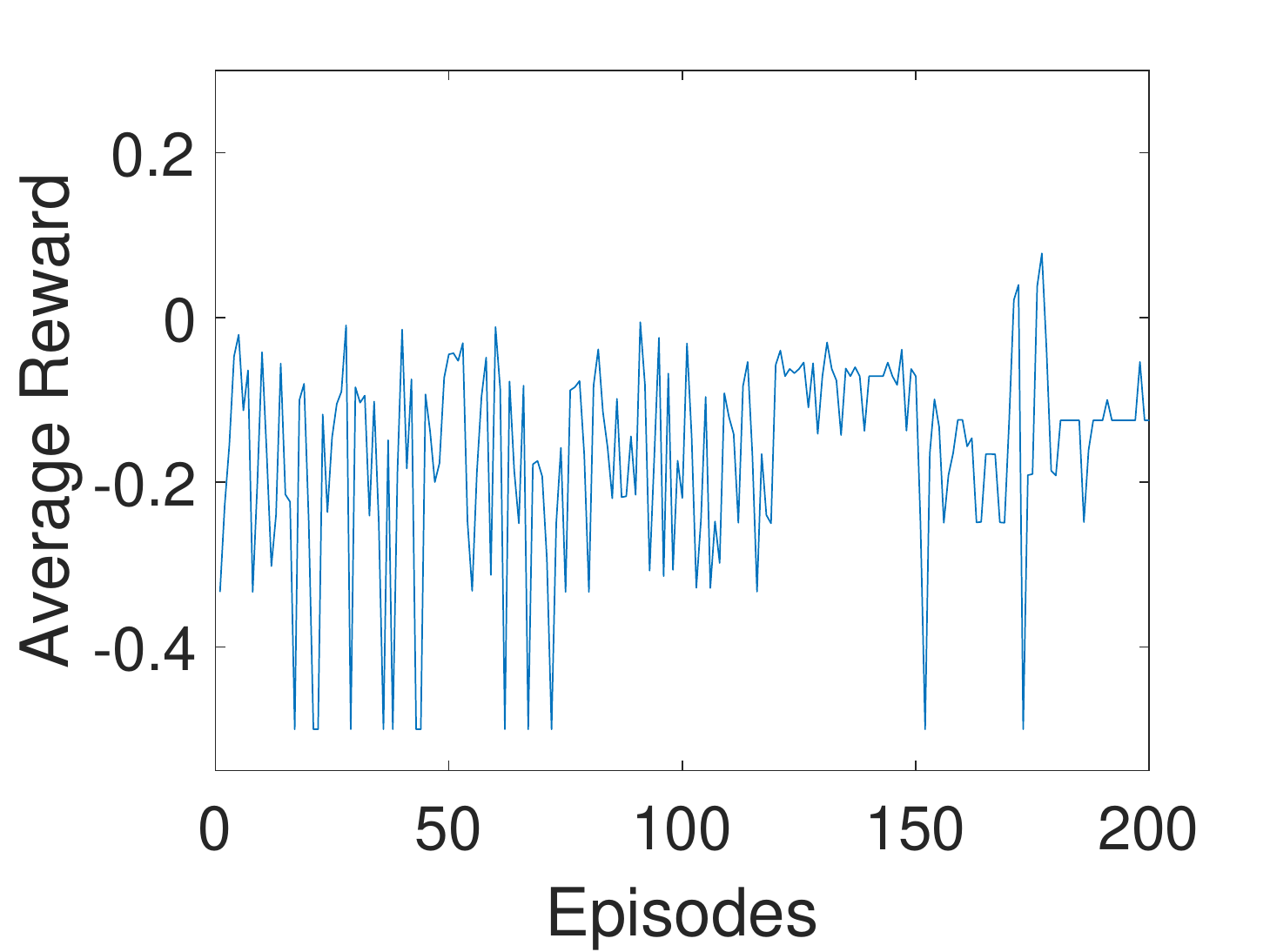}
  \end{minipage}
  \begin{minipage}{0.49\columnwidth}
    \centering
    \includegraphics[width=\columnwidth]{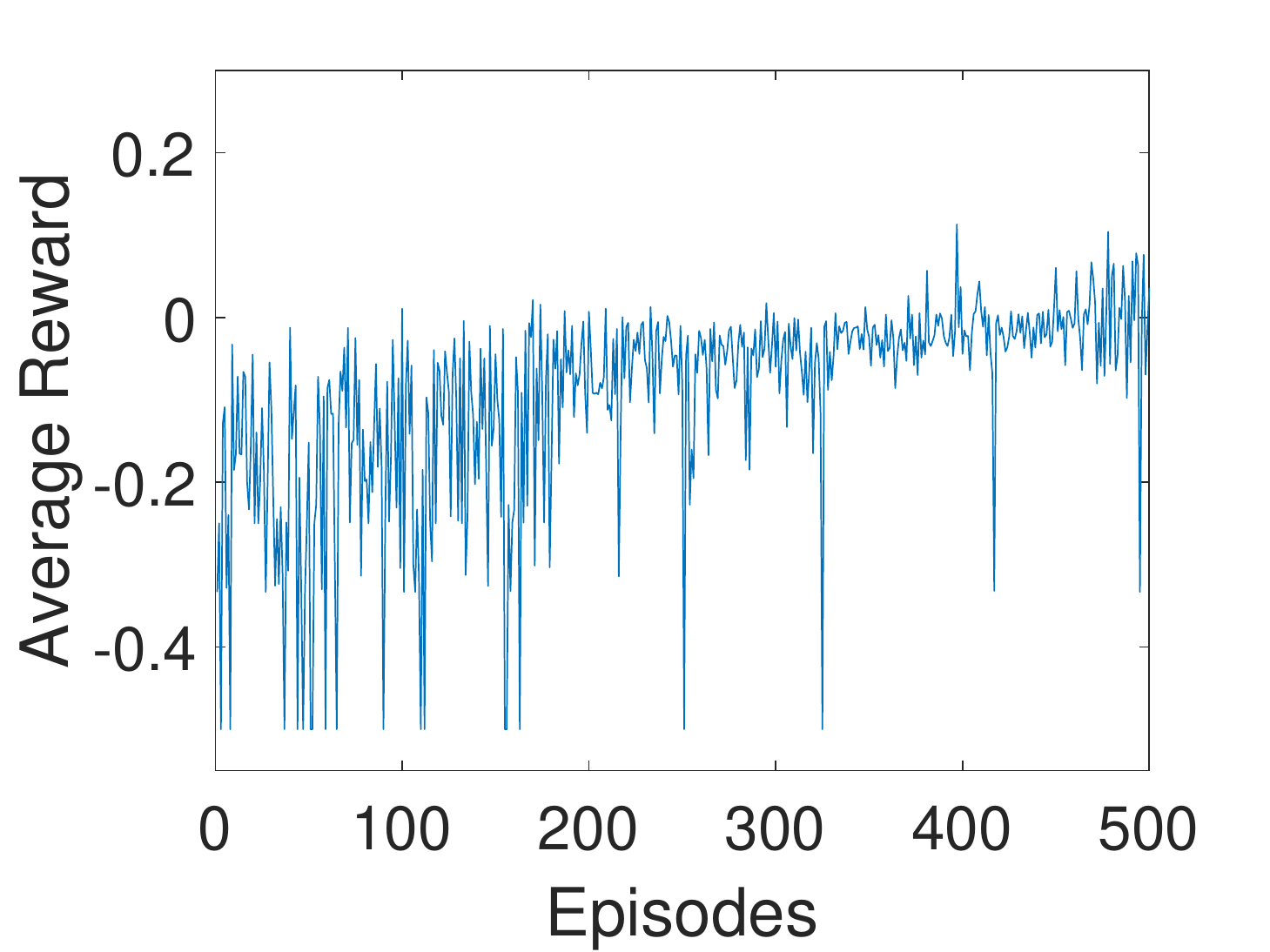}
  \end{minipage}
  \begin{minipage}{0.49\columnwidth}
    \centering
    \includegraphics[width=\columnwidth]{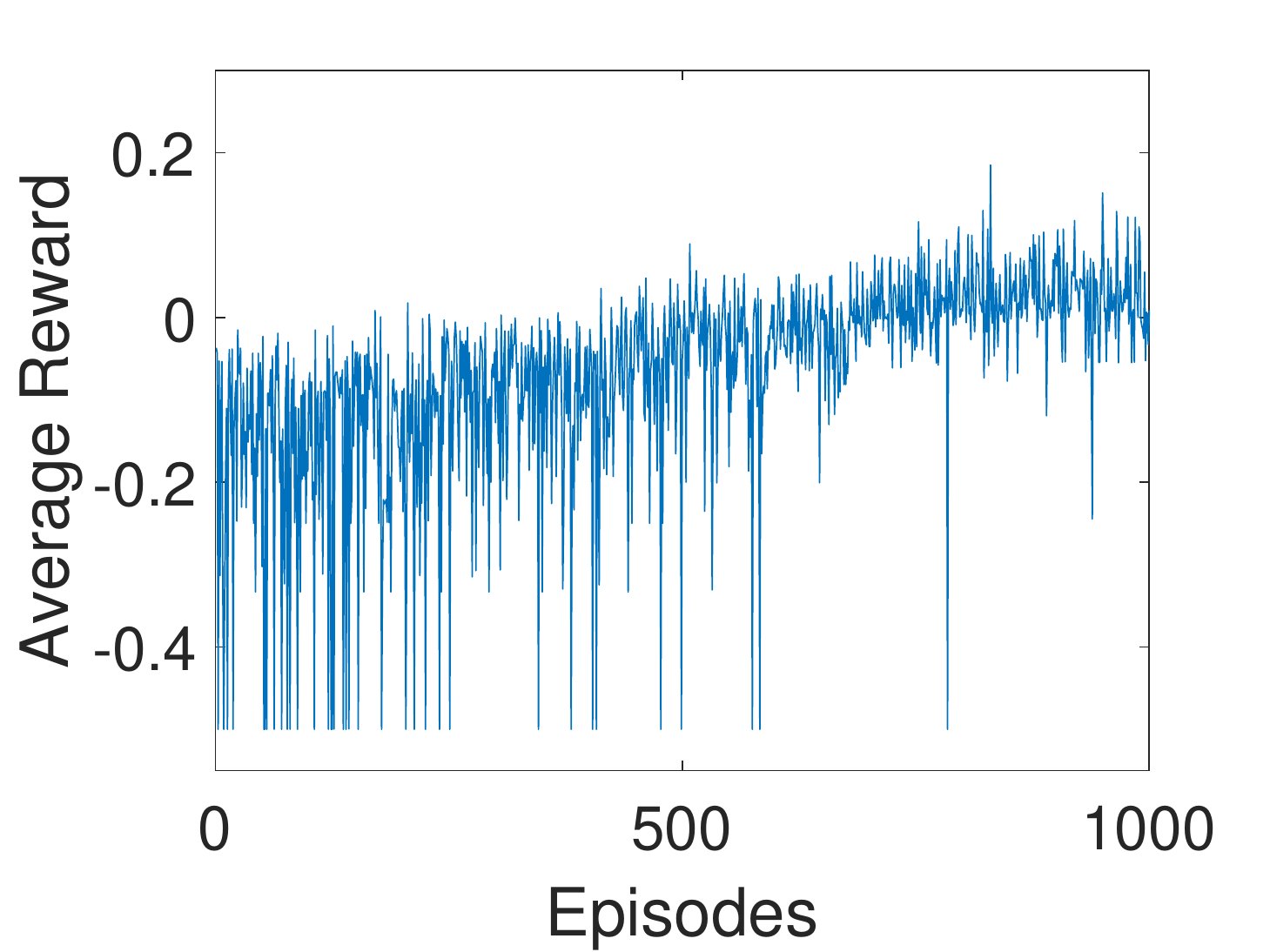}
  \end{minipage}
  \begin{minipage}{0.49\columnwidth}
    \centering
    \includegraphics[width=\columnwidth]{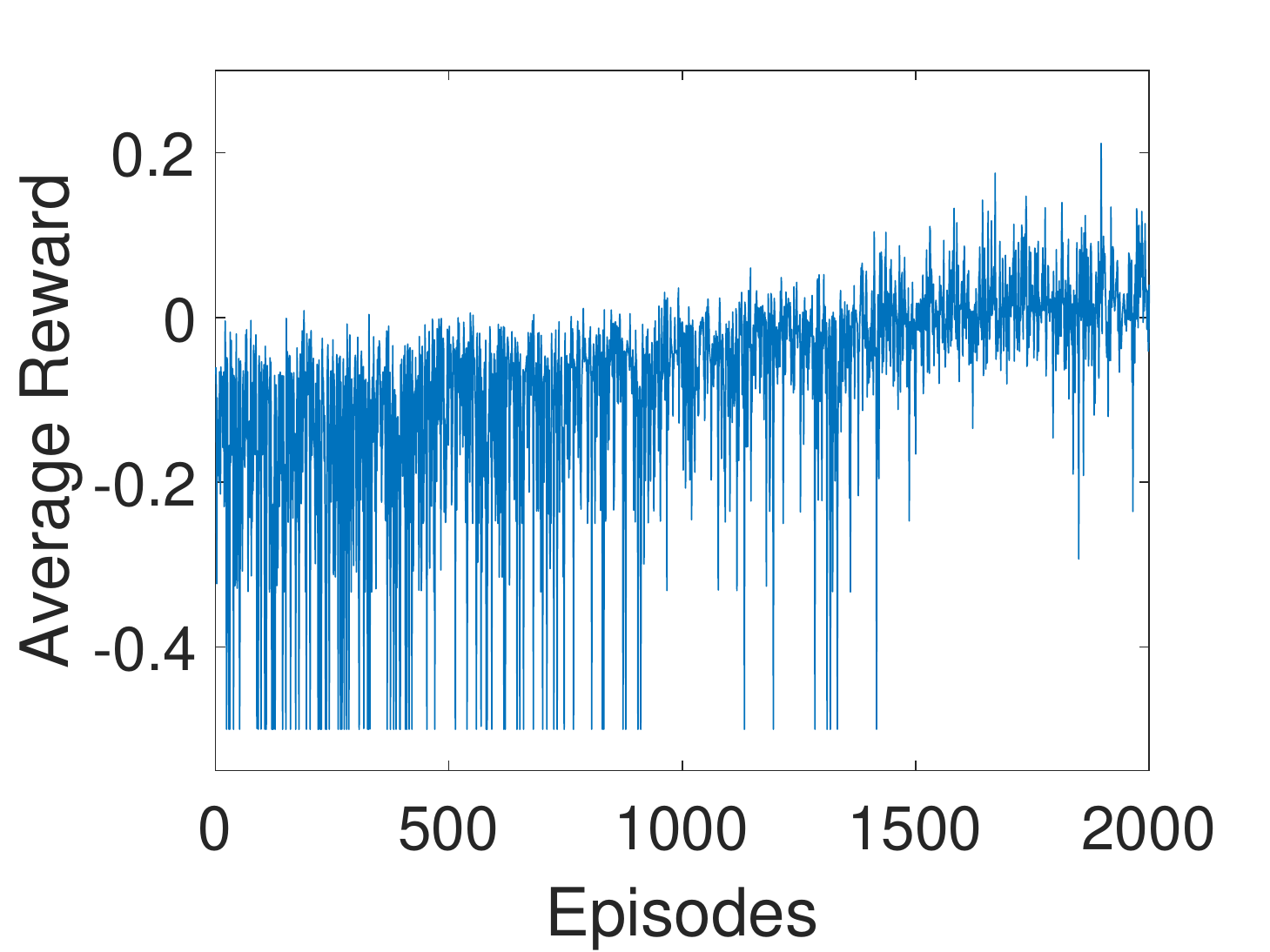}
  \end{minipage}
  \caption{Average rewards for five different training sessions with maximum numbers of episodes of 100, 200, 500, 1000, 2000.}
  \label{average_rewards}
\end{figure}

The average rewards obtained for each session are depicted in Fig.~\ref{average_rewards}. The average reward increases with the increased maximum number of episodes since the agent has more chances to interact with the environment and learn more. During the first training with 100 episodes which takes around 1-2 hours, the agent can barely reach positive average reward values. Learning is not visible for this case. In the second training with 200 episodes, a similar trend is observed without a visible increment on average rewards. In the third case with 500 episodes, the reward increment starts to become visible but there are still relatively lower values towards the end. In the fourth and fifth cases with 1000 and 2000 episodes respectively, increasing reward trend is quite prominent. Final average rewards are higher in these cases, while the agent trained for 2000 episodes has slightly better performance. The maximum number of episodes is limited at 2000 because the training takes around 20-22 hours for this case, beyond which the main motivation of this work (a practical RL system with early convergence) would be lost.

\subsection{Testing in AirSim}
\begin{figure}[t!]
  \centering
  \includegraphics[width=0.9\columnwidth]{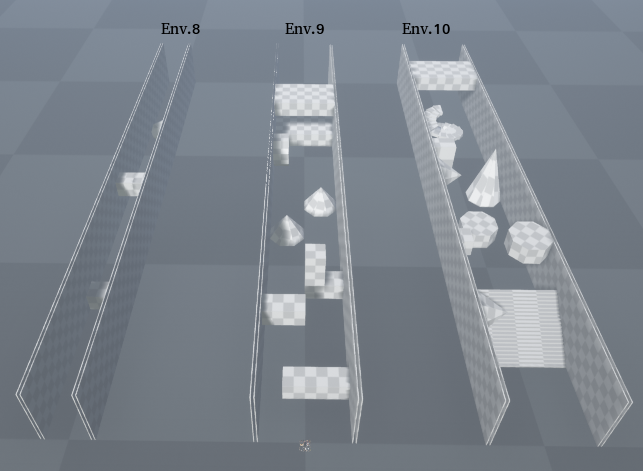}
  \caption{Test environments with obstacles of different shapes and sizes.}
  \label{airsim_test}
\end{figure}

In this section, the performance of the agent trained for 2000 episodes is evaluated through navigation from the start to the goal in ten different environments in AirSim. The first seven environments are the same as in training sessions, where the last three environments are previously unseen and they are depicted in Fig.~\ref{airsim_test}. While the former group examines whether the agent is competent on exploring a desirable policy within training environments, the latter group challenges the agent in terms of generalization to different environments. They consist of obstacles with different shapes and sizes, corridors with different widths but the same length\footnote{The tracks with the same straight length are considered for consistent benchmarking. However, the proposed method can easily be employed in different environments with different lengths of straight lines, a concatenation of straight lines, or turns. Once the intermediate goal points are given to create straight path segments as initial rough paths, it is trivial to utilize the method in these environments since the underlying RL system is designed with respect to the quadrotor's body frame.} of 60m.
\begin{table}[t!]
\renewcommand{\tabularxcolumn}[1]{>{\centering}m{#1}}
\caption{Evaluative test results in AirSim.}
\label{test_in_envs}
\centering
\begin{tabular}{ccccc}
\hline\noalign{\smallskip}
\multirow{2}{*}{\textbf{Env.$\#$}} & \textbf{Navigation} & \textbf{Navigation} & \multirow{2}{*}{\textbf{Crash}} & \textbf{Total} \\
& \textbf{distance (m)} & \textbf{time (s)} &  & \textbf{reward}\\ 
\noalign{\smallskip}\hline\noalign{\smallskip}
\multirow{5}{*}{\textbf{Env.1}} 
& 60.00 & 57 & N & 12.15\\
& 60.00 & 57 & N & 11.89 \\
& 60.00 & 57 & N & 10.94 \\
& 60.00 & 58 & N & 12.66 \\
& 60.00 & 57 & N & 11.51 \\
\hline\noalign{\smallskip}
\multirow{5}{*}{\textbf{Env.2}} 
& 60.00 & 58 & N & 11.93\\
& 60.00 & 57 & N & 12.49 \\
& 60.00 & 58 & N & 12.43 \\
& 60.00 & 57 & N & 12.28 \\
& 60.00 & 58 & N & 11.83 \\
\hline\noalign{\smallskip}
\multirow{5}{*}{\textbf{Env.3}} 
& 60.00 & 58 & N & 10.35\\
& 60.00 & 58 & N & 11.93 \\
& 60.00 & 57 & N & 12.16 \\
& 60.00 & 58 & N & 11.96 \\
& 60.00 & 58 & N & 12.14 \\
\hline\noalign{\smallskip}
\multirow{5}{*}{\textbf{Env.4}} 
& 60.00 & 79 & N & 3.82\\
& 60.00 & 71 & N & 4.44 \\
& 48.85 & 120 & N & 3.61 \\
& 29.96 & 42 & Y & 1.32 \\
& 28.15 & 33 & Y & 1.25 \\
\hline\noalign{\smallskip}
\multirow{5}{*}{\textbf{Env.5}} 
& 60.00 & 78 & N & 3.41\\
& 60.00 & 75 & N & 3.58 \\
& 60.00 & 71 & N & 3.53 \\
& 21.35 & 28 & Y & 0.51 \\
& 48.57 & 120 & N & 3.39 \\
\hline\noalign{\smallskip}
\multirow{5}{*}{\textbf{Env.6}} 
& 60.00 & 61 & N & 4.85\\
& 44.44 & 120 & N & 4.35 \\
& 60.00 & 70 & N & 4.46 \\
& 60.00 & 61 & N & 4.74 \\
& 60.00 & 61 & N & 4.81 \\
\hline\noalign{\smallskip}
\multirow{5}{*}{\textbf{Env.7}} 
& 50.42 & 120 & N & 3.26\\
& 41.51 & 58 & Y & 0.53 \\
& 15.90 & 77 & Y & 0.14 \\
& 60.00 & 95 & N & 3.27 \\
& 47.33 & 120 & N & 3.04 \\
\hline\noalign{\smallskip}
\multirow{5}{*}{\textbf{Env.8}} 
& 60.00 & 59 & N & 7.69\\
& 60.00 & 59 & N & 7.61 \\
& 45.47 & 44 & Y & 5.82 \\
& 45.39 & 45 & Y & 4.52 \\
& 60.00 & 58 & N & 9.01 \\
\hline\noalign{\smallskip}
\multirow{5}{*}{\textbf{Env.9}} 
& 60.00 & 57 & N & 9.24\\
& 60.00 & 58 & N & 9.31 \\
& 60.00 & 57 & N & 8.78 \\
& 60.00 & 57 & N & 8.24 \\
& 60.00 & 58 & N & 8.60 \\
\hline\noalign{\smallskip}
\multirow{5}{*}{\textbf{Env.10}} 
& 60.00 & 61 & N & 5.35\\
& 60.00 & 58 & N & 5.29 \\
& 60.00 & 61 & N & 4.95 \\
& 60.00 & 60 & N & 5.35 \\
& 60.00 & 60 & N & 5.88 \\
\hline
\end{tabular}
\end{table}

The agent is tested in each environment five times to have generalized evaluative results. Table~\ref{test_in_envs} states the values for reward, navigation distance, navigation time, and crash rate. Although the initial rough path in these environments (a straight path of 60m from the start to the goal) can be navigated within 60s in the obstacle-free configuration, the maximum number of time steps is set to 120 during the testing stage in order to account for the desired deviations from the path for obstacle avoidance. 

In the first three environments, the agent is able to learn a policy which is close to the optimal. Successful navigation from the start to the goal is observed for each trial in these relatively simple environments. The navigation time performance is also desirable as being close to 60s in 1m/s setting. If the common position controllers of the quadrotor would have been used without any local motion planning, they would have taken a similar amount of time for following the path from the start to the goal. The crash rate is 0$\%$ in these environments, even for the narrow corridor in Env. 3. Therefore, the agent's performance is favorable for the first group.

In the next four environments, the agent's performance is little degraded with the added complexity. Still, the agent successfully reaches the goal position in 67\% of the trials. In terms of safety, the agent completes 73\% of the trials without a crash in these dense environments. Moreover, there is no single head-on crash. All the crashes occur when the quadrotor moves towards sides or up-down without going forward and obstacles are out of sight of the front-facing camera. This limitation of the proposed method results from the lack of a history of depth images (which can be a part of the future work). On the other hand, there are some trials such as the third one in Env. 4 in which the agent cannot reach the goal position but does not take decisions which might yield a crash either. It just keeps its position around the same particular obstacle-free portion of the environments in order to maintain safety. Overall, the agent's performance is promising considering the density level of these environments. 

In the last three environments, the agent yields better performance as compared to the former group. It successfully reaches the goal position in 87\% of the trials. The other 13\% stands for two trials in Env. 8 in which there is a large sphere blocking the narrow corridor towards the end. The agent cannot avoid it fully and the landing gear touches to the sphere in these two trials. Apart from these two incidents, the agent is able to avoid obstacles of different shapes and sizes successfully and reach the goal position within a decent navigation duration. Therefore, the agent's performance can be considered as desirable in terms of generalizing to previously unseen environments.

\subsection{Testing in Real Flights}
For the real flight testing, the agent trained for 2000 episodes is deployed directly on a DJI F330 Quadrotor equipped with the flight controller PX4 FMU, the companion computer Nvidia Jetson TX2, and the extrecoptive sensor Intel RealSense D435. All the planning codes are running on TX2 utilizing its GPU for DQN in PyTorch with Cuda option. D435 provides live depth images to DQN while PX4 is responsible for the low-level control. The odometry information is obtained using a motion capture system. 

\begin{figure}[t!]
\centering
\begin{minipage}{0.49\columnwidth}
\centering
\includegraphics[width=\columnwidth]{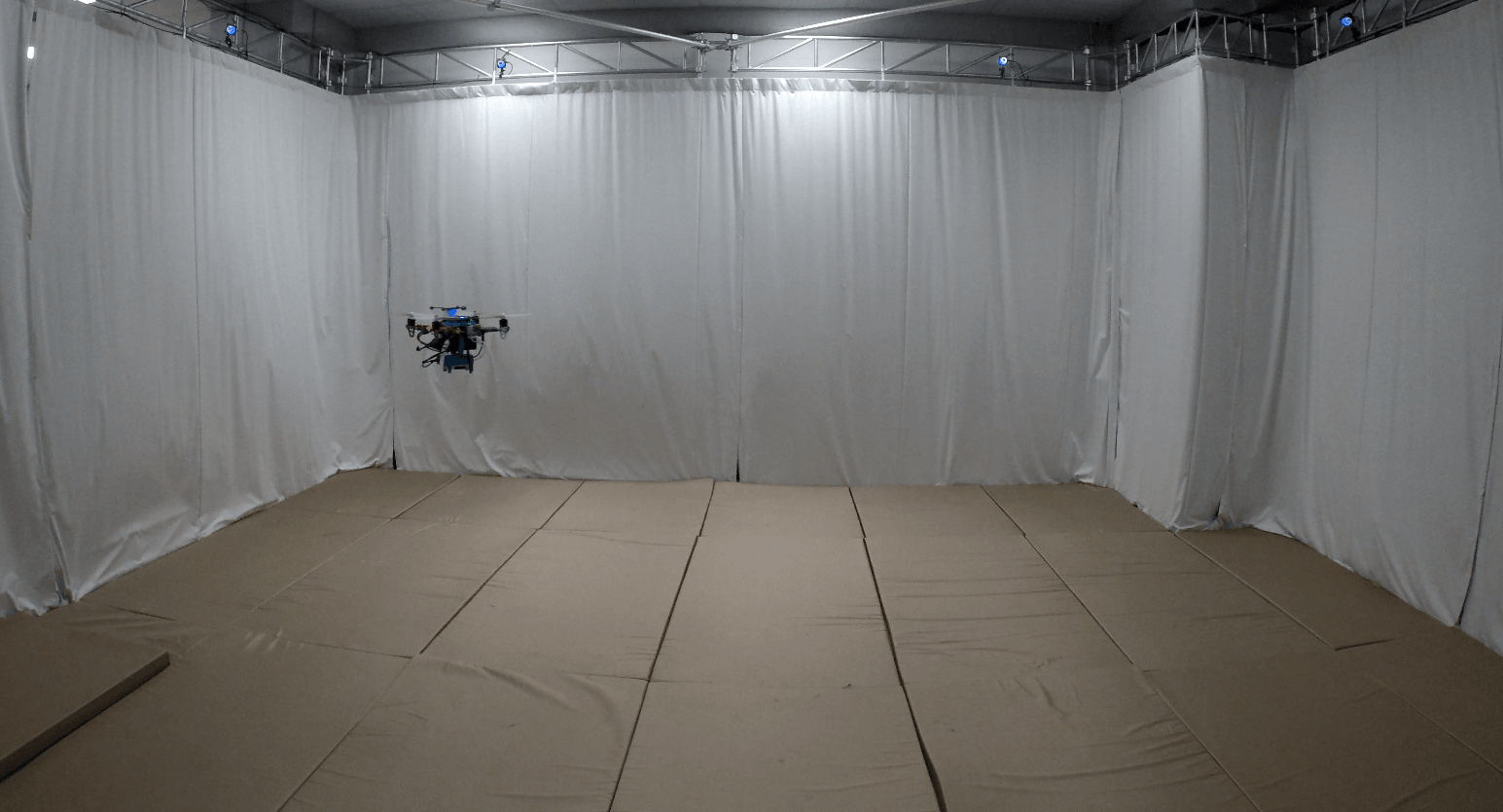}
\end{minipage}
\begin{minipage}{0.49\columnwidth}
\centering
\includegraphics[width=\columnwidth]{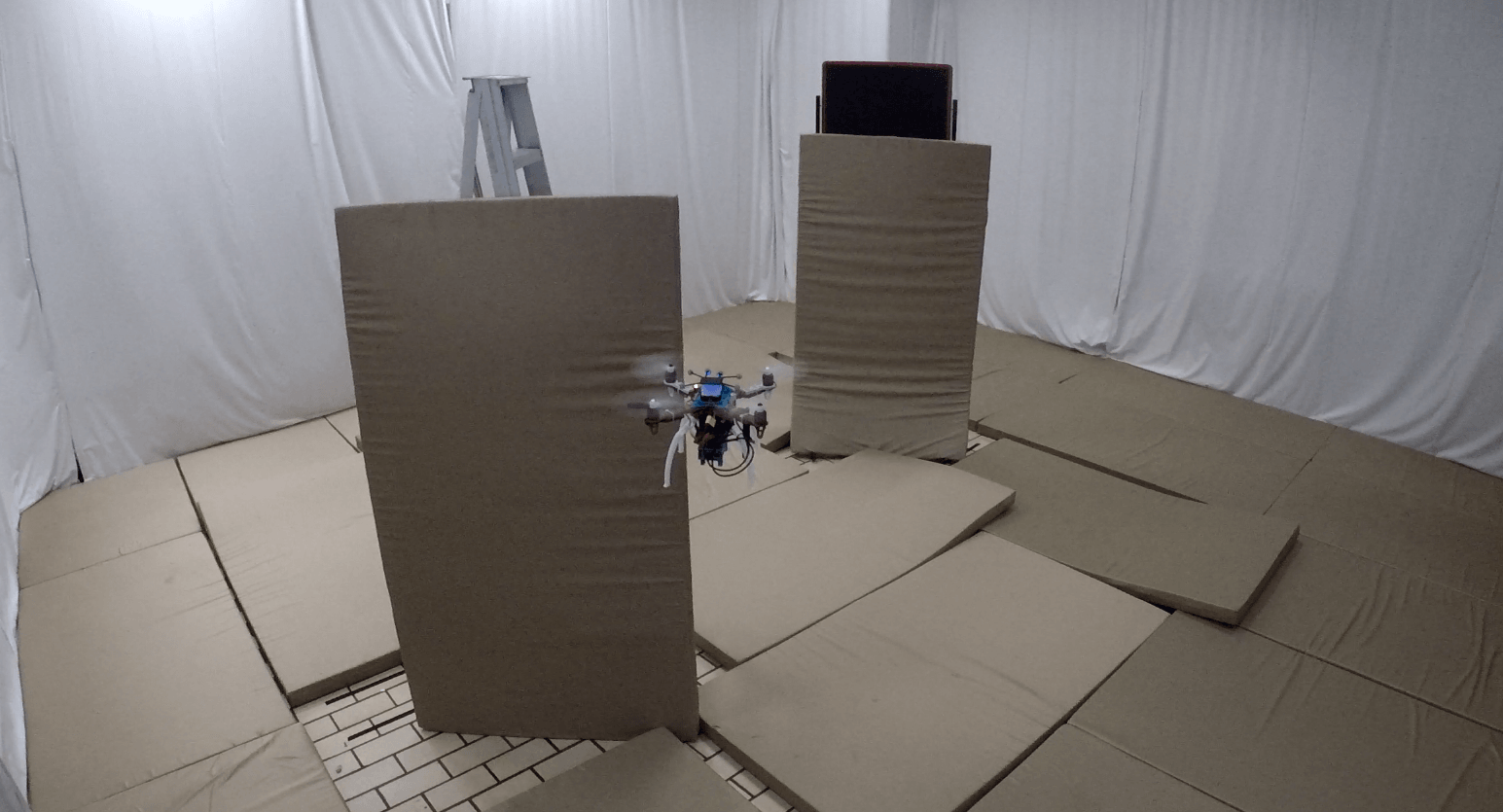}
\end{minipage}
\caption{Real flight environments. \textbf{Left:} Environment without obstacles. \textbf{Right:} Environment with two large rectangular shape obstacles.}
\label{real_trajectory_results}
\end{figure}

Two different environments are created for the real flights: one as obstacle-free and the other one with two large rectangular shape obstacles (Fig.~\ref{real_trajectory_results}). Since the lab space is limited to roughly 3m by 3m area, the initial path of 3.5m is set diagonally. In order not to cause dangerous movements of the quadrotor, the approximate speed of the vehicle is considered as 0.5m/s. Accordingly, the length of the motion primitives is decreased to a maximum movement of 0.5m in each axis. The maximum number of time steps are set to 30 in order to account for both the desired deviations from the initial path and the decreased length of the primitives. The rest of the parameters are exactly the same as in AirSim. Table~\ref{real_test_in_envs} summarizes the real flight test results.

In the first environment without obstacles, the agent's performance is desirable. There is no single crash and the agent reaches to the goal point with 100\% success out of five trials as can be seen in Table~\ref{real_test_in_envs}. The navigation time is relatively high as compared to the results in AirSim though. This is due to the motion primitive selections which cause deviations from the initial path generally in z axis and the control ability of the quadrotor in the real flights. Regarding the former, since the agent is getting the depth image from D435 which yields relatively noisy data as compared to AirSim, the agent's task is more difficult in the real flights. It is supposed to conduct sufficient reasoning by handling these relatively unclear images implicitly. As regards to the latter, the PX4's position controller with default parameters for DJI F330 Quadrotor is used in order to execute the motion primitive sequences governed by DQN. As compared to the ideal conditions in AirSim in which perfect odometry information without any delay as well as a precise controller to execute the high-level commands are available without considering any motor-ESC-propeller efficiency issues, the real flight control ability is obviously less. Still, the agent yields adequate end-to-end reasoning and completes the task with 100\% accuracy in the first environment.

In the second environment with two large rectangular shape obstacles, the agent's performance is relatively downgraded in terms of navigation to the goal. It cannot reach the goal position out of five trials. The furthest navigation point on the initial rough path is recorded as 1.55m in the third trial as can be seen in Table~\ref{real_test_in_envs}. On the other hand, the agent yields safe flights for 80\% of the trials in this environment. Only in the second trial, it crashes into the first obstacle. Again, it is not a head-on crash, the propeller just touches the obstacle from the side when the obstacle is out of sight of D435. In the other four trials, the agent avoids the first obstacle but then maintains its location in obstacle-free portions of the environment. Therefore, the agent's response in this environment can be considered as conservative. 

\begin{table}[t!]
\renewcommand{\tabularxcolumn}[1]{>{\centering}m{#1}}
\caption{Evaluative test results in real flights.}
\label{real_test_in_envs}
\centering
\begin{tabular}{ccccc}
\hline\noalign{\smallskip}
\multirow{2}{*}{\textbf{Env.$\#$}} & \textbf{Navigation} & \textbf{Navigation} & \multirow{2}{*}{\textbf{Crash}} & \textbf{Total} \\
& \textbf{distance (m)} & \textbf{time (s)} & & \textbf{reward} \\ 
\hline\noalign{\smallskip}
\multirow{5}{*}{\textbf{Env.1}} 
& 3.50 & 21 & N & 0.44\\
& 3.50 & 18 & N & 0.46 \\
& 3.50 & 21 & N & 0.44 \\
& 3.50 & 19 & N & 0.43 \\
& 3.50 & 22 & N & 0.43 \\
\hline\noalign{\smallskip}
\multirow{5}{*}{\textbf{Env.2}} 
& 0.65 & 30 & N & 0.34\\
& 1.03 & 15 & Y & -0.70 \\
& 1.55 & 30 & N & 0.32 \\
& 1.38 & 30 & N & 0.34 \\
& 0.80 & 30 & N & 0.37 \\
\hline
\end{tabular}
\end{table}

\section{DISCUSSION and FUTURE WORK}
As can be seen from the results, the proposed method demonstrates sufficient and promising performance for navigation of quadrotors in partially known environments. The agent yields fairly safe flights with collision-free flight percentage of 86\% over the trials in AirSim. This number is 90\% for real flights with relatively simpler environments. It reaches the goal point successfully during 88\% of these safe flights in AirSim, while this percentage is 56\% for real flights. Besides, the method caters for generalization to previously unseen environments which is particularly important to prove possible employability in a broad class of environments.

Overall, this work serves as a proof of concept for employing the proposed end-to-end RL-based motion planning system for quadrotor navigation in dense environments. It only includes specific case studies for brevity. Future work will include a comprehensive benchmarking study by considering a wider set of hyperparameters, different network structures, and more extensive experimental tests. Although the preliminary real flight test results demonstrate relatively safe navigation, the ultimate aim is to enhance the proposed method's performance on navigation to the goal, and present a simple and useful tool for the robotics community. 





\section*{ACKNOWLEDGMENT}
We would like to acknowledge the Air Lab members Rogerio Bonatti, Wenshan Wang, and Sebastian Scherer at Carnegie Mellon University, PA, USA for substantially helpful discussions. This work is financially supported by the Singapore Ministry of Education (RG185/17).

\addtolength{\textheight}{-12cm}
\bibliographystyle{IEEEtran}
\bibliography{RALbib}

\begin{thebibliography}{10}
\providecommand{\url}[1]{#1}
\csname url@rmstyle\endcsname
\providecommand{\newblock}{\relax}
\providecommand{\bibinfo}[2]{#2}
\providecommand\BIBentrySTDinterwordspacing{\spaceskip=0pt\relax}
\providecommand\BIBentryALTinterwordstretchfactor{4}
\providecommand\BIBentryALTinterwordspacing{\spaceskip=\fontdimen2\font plus
\BIBentryALTinterwordstretchfactor\fontdimen3\font minus
  \fontdimen4\font\relax}
\providecommand\BIBforeignlanguage[2]{{%
\expandafter\ifx\csname l@#1\endcsname\relax
\typeout{** WARNING: IEEEtran.bst: No hyphenation pattern has been}%
\typeout{** loaded for the language `#1'. Using the pattern for}%
\typeout{** the default language instead.}%
\else
\language=\csname l@#1\endcsname
\fi
#2}}

\bibitem{huang2017visual}
A.~S. Huang, A.~Bachrach, P.~Henry, M.~Krainin, D.~Maturana, D.~Fox, and
  N.~Roy, ``Visual odometry and mapping for autonomous flight using an rgb-d
  camera,'' in \emph{Robotics Research}.\hskip 1em plus 0.5em minus 0.4em\relax
  Springer, 2017, pp. 235--252.

\bibitem{stevens2018vision}
J.-L. Stevens and R.~Mahony, ``Vision based forward sensitive reactive control
  for a quadrotor vtol,'' in \emph{2018 IEEE/RSJ International Conference on
  Intelligent Robots and Systems (IROS)}.\hskip 1em plus 0.5em minus
  0.4em\relax IEEE, 2018, pp. 5232--5238.

\bibitem{sun2018robust}
K.~Sun, K.~Mohta, B.~Pfrommer, M.~Watterson, S.~Liu, Y.~Mulgaonkar, C.~J.
  Taylor, and V.~Kumar, ``Robust stereo visual inertial odometry for fast
  autonomous flight,'' \emph{IEEE Robotics and Automation Letters}, vol.~3,
  no.~2, pp. 965--972, 2018.

\bibitem{yang2019cubeslam}
S.~Yang and S.~Scherer, ``Cubeslam: Monocular 3-d object slam,'' \emph{IEEE
  Transactions on Robotics}, 2019.

\bibitem{camci2019planning}
E.~Camci and E.~Kayacan, ``Planning swift maneuvers of quadcopter using motion
  primitives explored by reinforcement learning,'' in \emph{2019 American
  Control Conference (ACC)}.\hskip 1em plus 0.5em minus 0.4em\relax IEEE, 2019,
  pp. 279--285.

\bibitem{morrell2018comparison}
B.~Morrell, R.~Thakker, G.~Merewether, R.~Reid, M.~Rigter, T.~Tzanetos, and
  G.~Chamitoff, ``Comparison of trajectory optimization algorithms for
  high-speed quadrotor flight near obstacles,'' \emph{IEEE Robotics and
  Automation Letters}, vol.~3, no.~4, pp. 4399--4406, 2018.

\bibitem{lai2018optimal}
S.~Lai, M.~Lan, and B.~M. Chen, ``Optimal constrained trajectory generation for
  quadrotors through smoothing splines,'' in \emph{2018 IEEE/RSJ International
  Conference on Intelligent Robots and Systems (IROS)}.\hskip 1em plus 0.5em
  minus 0.4em\relax IEEE, 2018, pp. 4743--4750.

\bibitem{camci2019learning}
E.~Camci and E.~Kayacan, ``Learning motion primitives for planning swift
  maneuvers of quadrotor,'' \emph{Autonomous Robots}, vol.~43, no.~7, pp.
  1733--1745, 2019.

\bibitem{greeff2018flatness}
M.~Greeff and A.~P. Schoellig, ``Flatness-based model predictive control for
  quadrotor trajectory tracking,'' in \emph{2018 IEEE/RSJ International
  Conference on Intelligent Robots and Systems (IROS)}.\hskip 1em plus 0.5em
  minus 0.4em\relax IEEE, 2018, pp. 6740--6745.

\bibitem{tal2018accurate}
E.~Tal and S.~Karaman, ``Accurate tracking of aggressive quadrotor trajectories
  using incremental nonlinear dynamic inversion and differential flatness,'' in
  \emph{2018 IEEE Conference on Decision and Control (CDC)}.\hskip 1em plus
  0.5em minus 0.4em\relax IEEE, 2018, pp. 4282--4288.

\bibitem{tang2018learning}
G.~Tang, W.~Sun, and K.~Hauser, ``Learning trajectories for real-time optimal
  control of quadrotors,'' in \emph{2018 IEEE/RSJ International Conference on
  Intelligent Robots and Systems (IROS)}.\hskip 1em plus 0.5em minus
  0.4em\relax IEEE, 2018, pp. 3620--3625.

\bibitem{mehndiratta2018automated}
M.~Mehndiratta, E.~Camci, and E.~Kayacan, ``Automated tuning of nonlinear model
  predictive controller by reinforcement learning,'' in \emph{2018 IEEE/RSJ
  International Conference on Intelligent Robots and Systems (IROS)}.\hskip 1em
  plus 0.5em minus 0.4em\relax IEEE, 2018, pp. 3016--3021.

\bibitem{sampedro2018image}
C.~Sampedro, A.~Rodriguez-Ramos, I.~Gil, L.~Mejias, and P.~Campoy,
  ``Image-based visual servoing controller for multirotor aerial robots using
  deep reinforcement learning,'' in \emph{2018 IEEE/RSJ International
  Conference on Intelligent Robots and Systems (IROS)}.\hskip 1em plus 0.5em
  minus 0.4em\relax IEEE, 2018, pp. 979--986.

\bibitem{everett2018motion}
M.~Everett, Y.~F. Chen, and J.~P. How, ``Motion planning among dynamic,
  decision-making agents with deep reinforcement learning,'' in \emph{2018
  IEEE/RSJ International Conference on Intelligent Robots and Systems
  (IROS)}.\hskip 1em plus 0.5em minus 0.4em\relax IEEE, 2018, pp. 3052--3059.

\bibitem{do2018learning}
C.~Do, C.~Gordillo, and W.~Burgard, ``Learning to pour using deep deterministic
  policy gradients,'' in \emph{2018 IEEE/RSJ International Conference on
  Intelligent Robots and Systems (IROS)}.\hskip 1em plus 0.5em minus
  0.4em\relax IEEE, 2018, pp. 3074--3079.

\bibitem{gschwindt2019can}
M.~Gschwindt, E.~Camci, R.~Bonatti, W.~Wang, E.~Kayacan, and S.~Scherer, ``Can
  a robot become a movie director? learning artistic principles for aerial
  cinematography,'' \emph{arXiv preprint arXiv:1904.02579}, 2019.

\bibitem{mnih2015human}
V.~Mnih, K.~Kavukcuoglu, D.~Silver, A.~A. Rusu, J.~Veness, M.~G. Bellemare,
  A.~Graves, M.~Riedmiller, A.~K. Fidjeland, G.~Ostrovski, \emph{et~al.},
  ``Human-level control through deep reinforcement learning,'' \emph{Nature},
  vol. 518, no. 7540, p. 529, 2015.

\bibitem{sutton1998reinforcement}
R.~S. Sutton and A.~G. Barto, \emph{Reinforcement learning: An
  introduction}.\hskip 1em plus 0.5em minus 0.4em\relax MIT press Cambridge,
  1998, vol.~1, no.~1.

\bibitem{huber1964robust}
P.~J. Huber \emph{et~al.}, ``Robust estimation of a location parameter,''
  \emph{The annals of mathematical statistics}, vol.~35, no.~1, pp. 73--101,
  1964.

\bibitem{kingma2014adam}
D.~P. Kingma and J.~Ba, ``Adam: A method for stochastic optimization,''
  \emph{arXiv preprint arXiv:1412.6980}, 2014.

\end{thebibliography}

\end{document}